\documentclass[lettersize,journal]{IEEEtran}
\usepackage{amsmath,amsfonts}

\usepackage{array}
\usepackage[caption=false,font=normalsize,labelfont=sf,textfont=sf]{subfig}
\usepackage{textcomp}
\usepackage{stfloats}
\usepackage{url}
\usepackage{verbatim}
\usepackage{graphicx}
\usepackage{cite}

\usepackage{longtable}
\usepackage{booktabs}
\usepackage{multirow}
\usepackage{makecell}
\usepackage{diagbox}

\usepackage{enumitem}

\usepackage{xcolor}

\usepackage[linesnumbered, ruled,vlined]{algorithm2e}
\SetAlFnt{\footnotesize} 

\SetCommentSty{mycommfont}

\usepackage{orcidlink}

\begin{document}

\title{Automated Keywords Generation Using LLMs in a High-Volume Data Virtualization Service}

\author{
    Saiful Khan\,\orcidlink{0000-0002-6796-5670}$^{1}$,
    Yiwen Xing\,\orcidlink{0000-0003-1521-6616}$^{2}$, 
    Joyraj Chakraborty\,\orcidlink{1111-2222-3333-4444}$^{2}$,
    Afrah Farea\,\orcidlink{0000-0003-4412-5377}$^{2}$,\\ 
    Philip Beaucamp\,\orcidlink{1111-2222-3333-4444}$^{2}$, 
    Samuel Jackson\,\orcidlink{0000-0001-5301-5095}$^{3}$, 
    and Min Chen\,\orcidlink{0000-0001-5320-5729}$^{2}$
    \thanks{$^{1}$UK Research and Innovation (UKRI), STFC, OX11 0QX, UK.}
    \thanks{$^{2}$University of Oxford, Oxford OX1 3QG, UK.}
    \thanks{$^{3}$United Kingdom Atomic Energy Authority (UKAEA), OX14 3EB, UK.}
}



\maketitle

\begin{abstract}
In modern data infrastructure, a data virtualization service (DVS) allows users to create many versions of source datasets for different processing tasks. For example, a multivariate or multi-modality dataset can be used for different machine learning tasks (e.g., classification, detection, segmentation, forecasting, and so on). Data selection, participation, integration, and preprocessing in different tasks result in many virtual datasets \cite{Khan:2025:SCC}. Meanwhile, different users need to discover existing virtual datasets frequently and easily to avoid duplicate efforts. To enable an effective search engine in a DVS, it is desirable to automatically generate keywords for each virtual dataset. In this work, we present a solution for generating keywords for virtual datasets using large language models (LLMs).
We propose five strategies for selecting different collections of text data for processing by the LLMs. We applied the five strategies to virtual datasets created in five application case studies with different source datasets. We conducted experiments with three LLM generators and two embedding techniques.
The experiments show that the performances of the five strategies depend on the quality of the text descriptions. The results help us adopt a multi-strategy implementation of a new keyword generation service (KGS) as part of an extended DVS.
The work demonstrates that the capability of automated keyword generation is indispensable for a high-volume DVS that is dataset-rich and frequently accessed. 
\end{abstract}


\section{Introduction}
\label{sec:Intro}
\IEEEPARstart{D}{ata virtualization} has been a common service across a variety of computational infrastructures, including but not limited to data warehouses, cloud computing, and virtual machines. With the rapid advancement of artificial intelligence (AI) and machine learning (ML), datasets used in AI/ML development are frequently transformed into different forms, e.g., filtering, merging, enriching, and pre-processing, which are then repeatedly accessed by many AI/ML developers. To reduce excessive resource waste from storing the same information in different physical forms, contemporary data virtualization services (DVSs) have emerged in which transformed datasets are stored as virtual datasets (e.g., \cite{Khan:2025:SCC}). 

In a DVS, a large number of virtual datasets are available to different AI/ML developers. One of their daily tasks is to search for existing virtual datasets before creating new virtual datasets. Such tasks may be performed by users who wish to be inspired by existing development activities while avoiding duplicated effort, and may also be performed by the DVS to identify risks of duplication and prompt a user (who is about to save a new virtual dataset) with such risks. Therefore, a DVS maintains a set of keywords for each virtual dataset as part of its metadata. This has created a nontrivial challenge about how to create these keyword sets in a cost-effective and consistent manner. If a DVS relies on users to define keywords when creating a new virtual dataset, discontent, slackness, and inconsistency would be inevitable. Meanwhile, poor-quality keyword sets would affect users' experience using the search facility in a DVS, as well as their ability to avoid duplicates.

In this paper, we report on our investigation into the feasibility of using large language models (LLMs) to generate a keyword set for every virtual dataset by processing its metadata and users' comments automatically. In particular, our contributions include:
\begin{enumerate}[leftmargin=1em, labelsep=0.3em]
    \item[(a)] We present an enhanced design of a data virtualization service (DVS), which provides automatic keyword generation with useful raw data, including (i) metadata captured by user interfaces, (ii) relationships among different datasets, and (iii) users' comments about datasets. The DVS generates and maintains (i) and (ii) automatically.%
    \item[(b)] We present a novel \emph{keyword generation service} (KGS) in the context of service computing in general and DVS in particular, and utilize LLMs to generate keywords automatically for numerous datasets in a high-volume DVS.%
    \item[(c)] We report a series of experiments to answer a range of research questions about the technology's sensitivity to (i) different technical components and parameters of LLMs, (ii) availability, quantity, and quality of users' comments, (iii) uses of keywords of parent datasets, and application contexts of the datasets.%
\end{enumerate}

Together, we demonstrate the necessity and feasibility of automated keyword generation in a high-volume DVS. The experimental results show that the LLM techniques can be optimized to obtain strong performance, while different types of sensitivity can be significantly reduced when the DVS provides the KGS with multiple types of raw data. Therefore, LLMs-enabled keyword generation is most likely applicable to other aspects of infrastructure-based service computing.
We provide open-source code and data at \url{https://github.com/saifulkhan/automated-keyword-generation-with-llm}.

\section{Related Work}

\vspace{1mm}\noindent\textbf{Automatic Metadata Generation.}
Dataset discoverability depends largely on metadata. Gebru et al.~\cite{Gebru2021DatasheetsDatasets} recommend documenting dataset composition, motivation, collection, and uses. Chapman et al.~\cite{Chapman2020DatasetSurvey} argue that data provide value only when users can find, access, and interpret them, making metadata quality central to discovery infrastructure. 
Akhtar et al.~\cite{Akhtar2024Croissant} introduce Croissant, a metadata format that standardizes descriptions of machine-learning dataset structure, resources, and intended use. Croissant represents and exchanges rather than generates metadata, complementing automatic metadata creation.

Related work automates metadata generation for structured data. Sherlock~\cite{Hulsebos2019Sherlock} uses deep learning on the VizNet corpus to detect 78 semantic types, outperforming dictionary- and regex-based methods.  
DODUO~\cite{Suhara2022AnnotatingModels} used LLMs and table context to annotate column types and relations, improving quality with as few as eight tokens per column. Chen et al.~\cite{Chen2019LearningData} combine table context, inter-column semantics, and knowledge-base lookups for accurate, transferable column-type prediction. Beyond fixed-label sets, Meng et al.'s~\cite{Meng2017DeepGeneration} encoder--decoder architecture produces free-form keyphrases, including phrases absent from the source text. 

Kong et al.~\cite{Kong2023PromptRank} propose PromptRank, an unsupervised method that uses a prompted encoder--decoder model to rank candidate keyphrases by document relevance.
Skluzacek et al.~\cite{Skluzacek2022AutomatedOpportunities} present a repository extraction system that improves scientific-data findability with little human involvement. 
AutoDDG~\cite{Zhang2025AutoDDG} uses LLMs to describe tabular datasets and improve retrieval but ignores the transformation history of derived data.
Hayashi et al.~\cite{Hayashi2024MetadataRAG} combine metadata, vector retrieval, and retrieval-augmented generation to find related datasets and estimate tags, assuming metadata already exists.

\vspace{1mm}\noindent\textbf{LLM in Service Oriented Computing.}
Within the service~\cite{Khan:2022:TSC,Khan:2022:JDIM,Khan:2023:IEEEAccess} life-cycle, a multitude of LLMs are being utilized. Kumara et al.~\cite{kumara2026llm4soc} present a systematic review of service-oriented computing tasks based on 64 papers, subdividing the field into service foundation, monitoring or management, composition, and design or development.
Several directions are relevant to producing semantic artifacts. Zhu et al.~\cite{zhu2025llmdoc} use LLMs to automate business-process documentation, improving ineffective, error-prone manual and rule-based approaches. Pesl et al.~\cite{pesl2025compositio} present the ``Compositio Prompto'' architecture, which uses task descriptions and service specifications to synthesize executable compositions without heavy formal models. 
Meflah et al.~\cite{meflah2026tosca} use LLMs to interpret natural-language intents and generate vendor-neutral TOSCA compositions. 
Nivon and Salaün~\cite{nivon2025bpmn} generate BPMN process models from text requirements, while Rong et al.~\cite{rong2025coldstart} combine semantic and topological signals for LLM-based cold-start service recommendations. 

These studies show that LLMs can generate semantic artifacts in service-oriented systems. However, prior work either annotates datasets without considering derived-data transformation history or addresses service life-cycle stages beyond dataset discoverability. We bridge this gap by embedding LLM-based keyword generation as a DVS sub-service, varying the provenance context supplied to the LLM, and quantifying keyword-quality and token-cost trade-offs across five heterogeneous case studies.

\section{Data Virtualization Service (DVS)}
\label{sec:DVS}
In a contemporary DVS, a virtual dataset does not store the ``real'' data that can be used directly. Instead, it stores information on how the ``real'' data can be \emph{transformed} from one or more preceding datasets. Such transforms include a variety of operations for data configuration (e.g., part-selection, merge, partition) and pre-processing (e.g., Fourier transformation, feature extraction, noise reduction) as well as remote access in traditional data virtualization. Typically, a DVS is equipped with a number of sub-services, including but not limited to,
\emph{Data Virtualization Manager},
\emph{Data Query Engine},
\emph{Data Processing Engine}, and
\emph{Transformations Repository}
\cite{Khan:2025:SCC}.

\vspace{1mm}\noindent\textbf{Infrastructural Support for Machine Learning.}
Research and development (R\&D) activities in ML can benefit tremendously from a DVS. ML researchers commonly conduct many experiments. For example, researchers developing ML models to denoise speech signals can create a variety of noisy speech signals by mixing a piece of clean speech data (i.e., ground-truth data) with different types of noise signals in different ratios \cite{reddy20_interspeech,11126969}. Given a clean speech dataset, one could easily create tens or even hundreds of different noisy speech datasets within a day. Such noisy speech datasets are ideally stored as virtual datasets, allowing them to be shared among researchers within a team over time, while maintaining data provenance and ensuring experimental consistency. Khan et al. reported six ML applications of a DVS \cite{Khan:2025:SCC}. It is not difficult to envision many other ML applications in which new datasets are created on a daily or weekly basis.

\begin{figure*}[t]
    \centering
    \includegraphics[width=.9\linewidth]{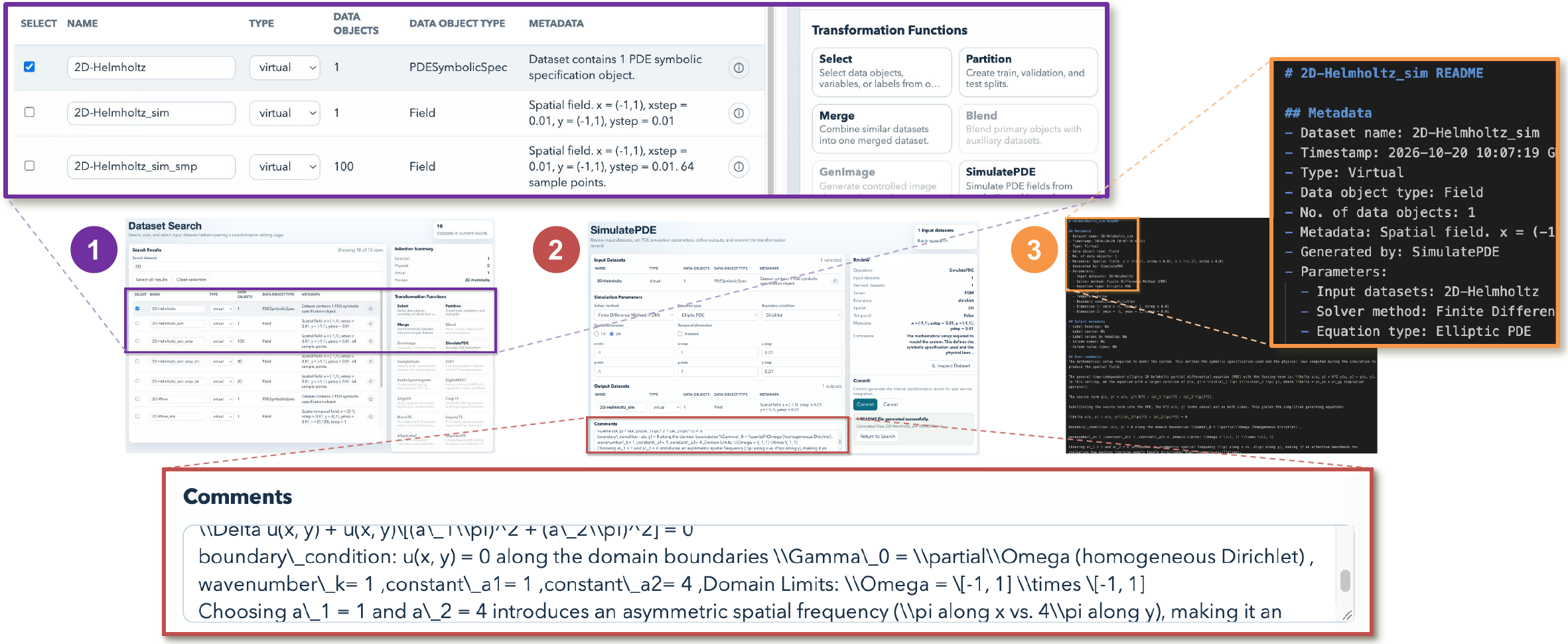}
    \caption{An example of creating a virtual dataset: (1) locating a parent dataset and invoking a data transformation; (2) using the corresponding UI to define parameters and create a virtual dataset; and (3) the README file of the virtual dataset captures a variety of information for creating a physical dataset as well as user comments. The information in the README file can be used to generate keywords for the dataset automatically.}
    \label{fig:UI}
\end{figure*}

Fig. \ref{fig:UI} shows an example of creating a virtual dataset. (1) A user locates a parent dataset, which contains $N$ sets of partial differential equations (PDEs). (2) The user invokes a data transformation \texttt{SimulatePDE}, uses the UI to define boundary conditions such that each set of PDEs is ready to be simulated, and creates a virtual dataset containing $N$ virtual fields as simulation results that can be produced on demand. (3) The README file of the virtual dataset captures the \emph{action documentation} as well as \emph{user comments}. In the README file, the name of the input dataset provides a link to the parent dataset, while the name of the transformation function also provides a link to the generic software documentation.

\begin{figure}[t]
    \centering
    \includegraphics[width=1\linewidth, trim=0cm 0cm 0cm 0cm, clip]{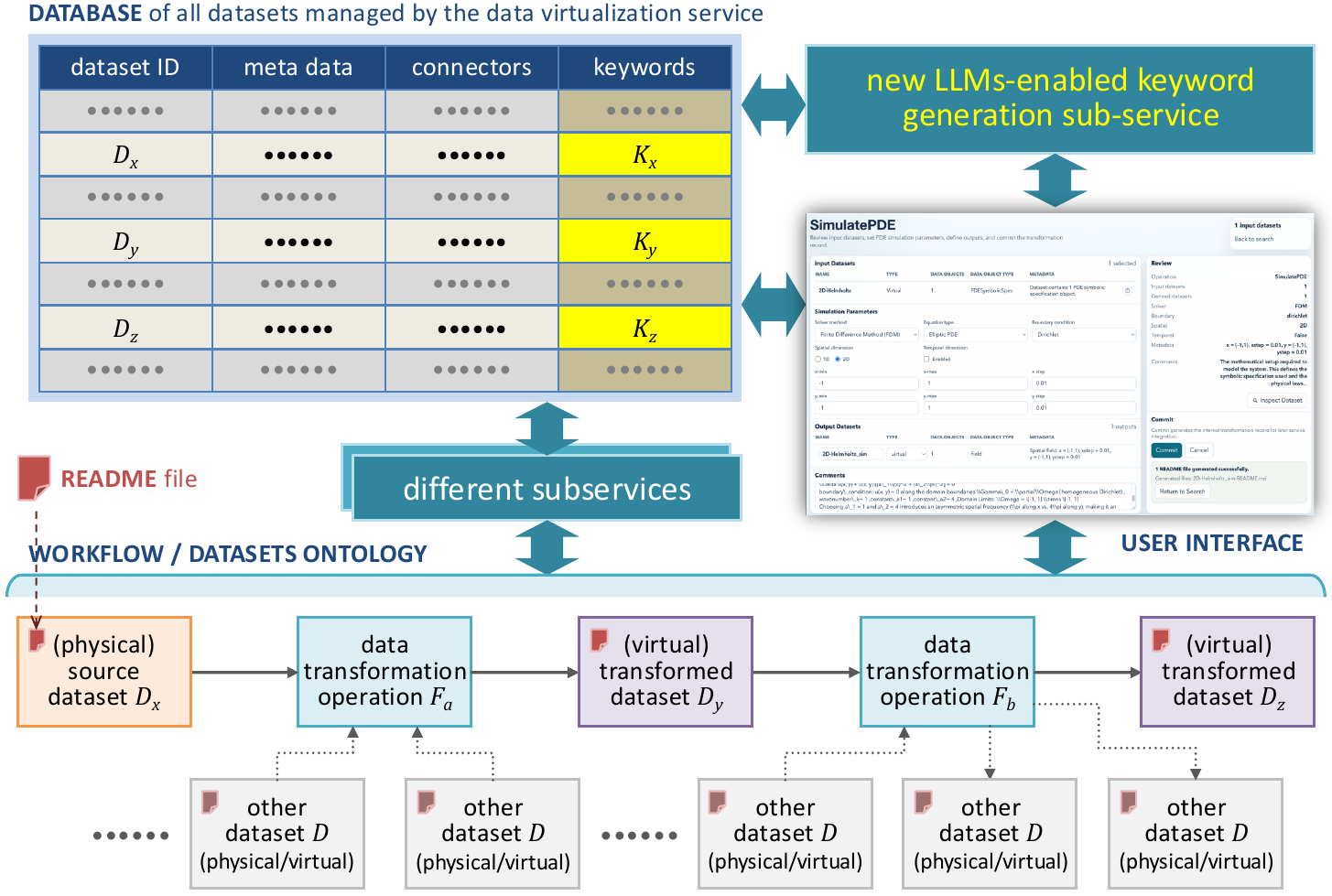}
    \caption{Keyword-related components in a DVS. \textbf{Bottom:} users generate numerous virtual datasets in their workflows for different ML tasks. These virtual datasets, together with their corresponding source datasets, are interrelated, and their relationships are conceptualized as a dataset ontology called DVG. \textbf{Top-Left:} The DVS manages a database about all physical and virtual datasets, including their unique IDs, their metadata, their relationships, as well as the keyword set of every dataset. \textbf{Top-Right:} When a user creates a virtual dataset, the user interface invokes the KGS that generates a keyword set for the newly created virtual dataset and stores it in the database.}
    \label{fig:Workflow}
\end{figure}

\vspace{1mm}\noindent\textbf{Data Virtualization Graph (DVG).}
Over time, a DVS accumulates numerous virtual datasets created by different ML researchers for various ML applications. These virtual datasets are interrelated, as illustrated in Fig.~\ref{fig:Workflow} (bottom). When a transformation function $F_a$ is used to transform a physical dataset $D_x$, together with some other datasets as input into a virtual dataset $D_y$, the directional relationships from $D_x$ (and other datasets input) to $F_a$ and then to $D_y$ must be maintained, so that when $D_y$ is needed by an ML task (e.g., training), its physical version can be calculated directly using $F_a$ and its input datasets. The clean and noisy speech datasets mentioned above are examples of $D_x$ and $D_y$. 
$D_y$ can be further transformed into other virtual datasets, e.g., by partitioning into training, validation, and testing datasets. The relationships among all datasets (virtual as well as physical) and the transformation functions concerned are conceptually an ontology~\cite{Khan:2016:AEI} in the form of a digraph that is always acyclic and bipartite, and can be connected or disconnected. This ontology is referred to as DVG.

\vspace{1mm}\noindent\textbf{Text Documentation.}
Every node in the DVG is either a dataset (physical or virtual) or a transformation function. We assume that all datasets and transformation functions are accompanied by some form of text documentation. Such texts may be created in the following ways:
\begin{enumerate}[leftmargin=1em, labelsep=0.3em]
    \item[a.] \emph{Imported from external sources} -- For example, an original physical dataset is typically imported from an external data repository, together with one or more text files describing the dataset.
    \item[b.] \emph{Essential and generic software documentation} -- Transformation functions in the \emph{Transformations Repository} of the DVS are expected to be accompanied by generic text descriptions of their functionality, parameters, inputs, and outputs. Similarly, user interfaces (UIs) that allow users to invoke transformation functions interactively to create virtual datasets are expected to be accompanied by generic text descriptions. 
    \item[c.] \emph{Action documentation} -- When a user creates a virtual dataset through the UI corresponding to the selected transformation function, the UI automatically records the information about this specific virtual dataset, e.g., the IDs of the input datasets and the parameter settings of the transformation function. 
    \item[d.] \emph{User comments} -- Through the UIs, users can enter additional comments when creating virtual datasets.
\end{enumerate}

To maintain the generality of the proposed solution without getting bogged down by implementation details of a specific DVS, we assume that every dataset (physical or virtual) has a README file or description that contains information of type (a) or types (c) and (d); type (b) information is centrally recorded as part of the \emph{Transformations Repository}. In our experiments to be detailed in Section \ref{sec:Experiments}, we refer to data of types (a), (b), and (c) collectively as ``metadata'', distinguishing them from type (d) ``user comments''. 

\vspace{1mm}\noindent\textbf{Need for Automated Keyword Generation.}
Enabling users to create virtual datasets brings many benefits, including cost savings in data storage, provenance of datasets and their usage, efficiency, and a collaborative culture through data sharing, and so on. It also makes \emph{search} an essential function in a DVS. As illustrated in Fig.~\ref{fig:Workflow} (top-left), a DVS manages a database about all datasets (physical and virtual), including their unique IDs, their metadata, and their relationships. To enable the \emph{Data Query Engine} (DQE) of the DVS, the database also maintains a set of keywords for every dataset. Although it may be possible to ask users to specify the keywords of a virtual dataset when it is created, doing so would collectively require considerable user effort; it would be difficult to maintain consistency among different users due to many factors. For instance, different users may have technical expertise with the datasets and how they could be used; they may be biased by their personal experience and their specific plan to use the data; their personal preference in choosing words; and their time constraints in optimizing a set of keywords.
 
\section{Keyword Generation Service (KGS) in the DVS}
\label{sec:KGS}

Keywords play an indispensable role in any large data infrastructure, where users need to search for datasets frequently. When a data infrastructure contains a large collection of datasets, manually specifying and updating all keywords becomes costly. With data virtualization, creating new datasets becomes very easy and inexpensive, but manually entering keywords for every dataset becomes burdensome and costly. 

We therefore introduce a technical solution that utilizes LLMs to generate a set of keywords from text descriptions related to a newly created virtual dataset. As shown in Fig.~\ref{fig:Workflow}, when a user creates a virtual dataset, the UI invokes the KGS that generates a keyword set for the newly-created virtual dataset and stores it in the database.

\begin{figure*}[t]
    \centering
    \includegraphics[width=0.8\linewidth]{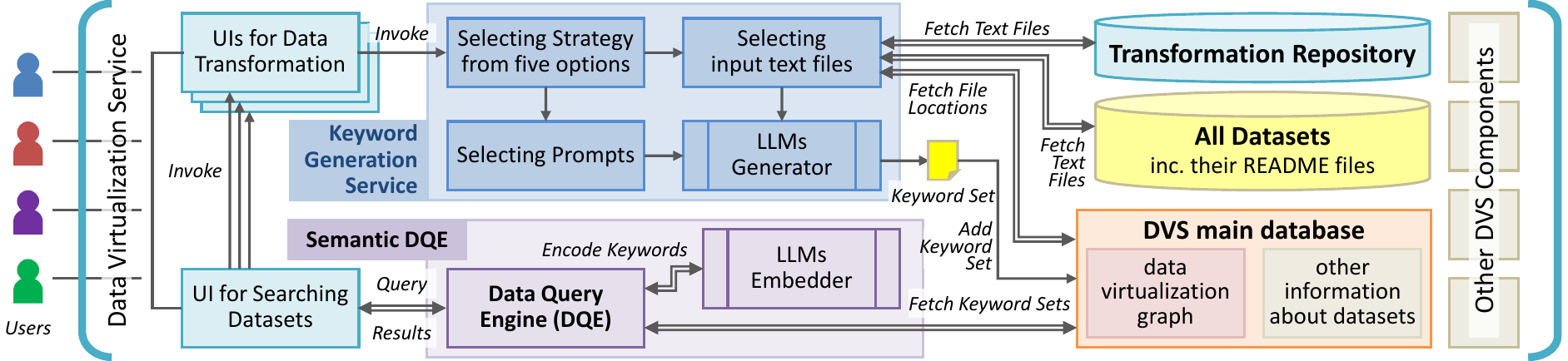}
    
    \caption{The KGS is a new component of an extended DVS~\cite{Khan:2025:SCC}. \textbf{Top}: It is invoked by one of the UIs for data transformation whenever a new virtual dataset is created. The generated keywords are stored in the DVG (data virtualization graph).  \textbf{Bottom}: The stored keywords are used by the DQE (data query engine) in the DVS. The DQE is extended with a new LLM-aided capability to match keywords semantically.}
    \label{fig:Architecture}
\end{figure*}

The KGS, in Fig.~\ref{fig:Architecture}, is implemented as an extension of the DVS~\cite{Khan:2025:SCC}. The DVS maintains two structures on which the KGS depends: a vector database that stores embeddings of metadata and keywords for all physical and virtual datasets, and a provenance graph modeled as a directed, labeled property graph. 
When a new virtual dataset is registered, the KGS is invoked in topological order over this graph. A provenance traversal component first resolves the node's immediate lineage, identifying its producing transformation and parent datasets. 

The assembled context is passed to a locally hosted open-weight generator model in our deployment, whose response is parsed, validated, and written back to the keywords property of the corresponding data-object node in the datasets database. 
Because keywords are stored as first-class node properties rather than derived at query time, the search and discovery interface can index them immediately. A vector database indexes the metadata and keywords as LLM-generated embeddings to support semantic search. The KGS therefore operates asynchronously alongside the existing DVS: virtualization operations remain unchanged, while every dataset in the high-volume service becomes discoverable through automatically generated semantically rich keywords.

\begin{figure*}[t]
    \centering
    \includegraphics[width=0.7\linewidth]{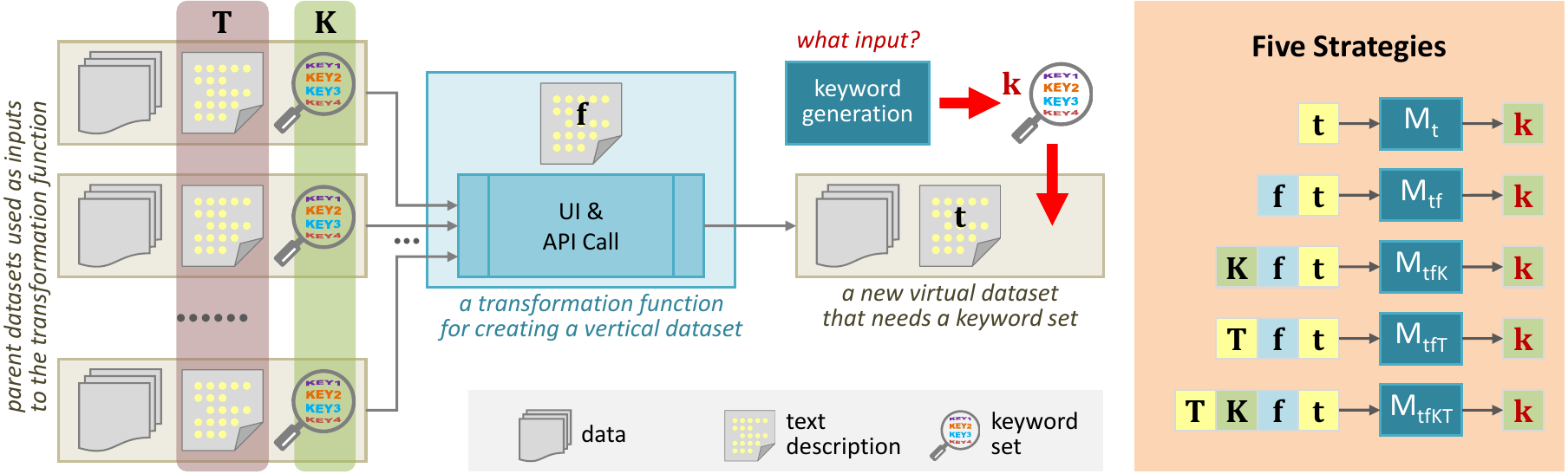}
    \caption{Five different keyword generation strategies studied in this work.}
    \label{fig:Methods}
\end{figure*}

\subsection{Core Techniques}
\label{sec:Techniques}
Given a piece of text, e.g., metadata or user comments, the KGS invokes a call to an LLM's API. Each LLM call includes two messages: a fixed generic system prompt and a user prompt specific to each dataset and control method. 

The ``system prompt'' tells the LLM generator models its role and the output format. It is defined consistently for all LLM APIs, control methods, and datasets. For our KGS, we use the system prompt: ``\emph{You are a metadata indexer. Generate concise search keywords (tags) for a dataset given its description and provenance context. Return ONLY a JSON array of 6 -- 10 short keyword strings (1 -- 4 words each), lowercase unless a proper noun, no explanations, no markdown.}''

The ``user prompt'' contains the raw data about one target dataset. It starts with the target's README text (abbreviated as \textbf{t}). Depending on the strategy of using additional data, we may add: the name or description of the transformation that produced the target (\textbf{f}), the keywords of the target's parents (\textbf{K}); and the README texts of the target's parents (\textbf{T}).
 
The user prompt ends with "\emph{Return the JSON array of keywords for the target dataset.}"

\vspace{1mm}\noindent\textbf{LLM Models.} 
Our implementation includes three open-source generator LLMs and two sentence-embedding models, as summarized in Table~\ref{tab:models}. Except for LLM gemma3:27b, which requires a datacenter-class GPU, all models have modest computational requirements and can run effectively on commodity personal computers. We ran all the experiments presented in Section~\ref{sec:Experiments} on an NVIDIA DGX cluster with two A100 GPUs (40\,GB VRAM). 

\begin{table}[h]
\caption{LLM models used in the implementation of our KGS.}
\label{tab:models}
\centering
\setlength{\tabcolsep}{4pt}
\begin{tabular}{llrrl}
    \toprule
    \textbf{Type} & \textbf{Model} & \textbf{Parameters} & \textbf{Size} & \textbf{Abbr.} \\
    \midrule
    Generator & gemma3:27b       & 27B   & 17 GB   & G27 \\
    Generator & llama3.2:3b      & 3.2B  & 2 GB    & L3 \\
    Generator & Mistral-7B       & 7.2B  & 13.4 GB & M7 \\
    Embedder  & Qwen3-Embedding-8B & 8B  & 4.7 GB  & Qwen \\
    Embedder  & all-MiniLM-L6-v2 & 22.7M & 91 MB   & MiniLM \\
    \bottomrule
\end{tabular}
\end{table}

\vspace{1mm}\noindent\textbf{Similarity Measures.}
When users of the DVS search for datasets, the \emph{Data Query Engine} invokes a similarity measure to compare keywords. Because of a huge number of datasets in a DVS and many users with different backgrounds, exact string matching is not always effective. We therefore use semantic similarity measures \cite{corley2005:sem-tex-sim:EMSEE,wu2024:kpeval:ACL,zhang2020:ICLR:bertscore}.

Let $K=\{k_1,\dots,k_n\}$ be the set of $n$ keywords of a dataset and $U=\{u_1,\dots,u_m\}$ the set of $m$ keywords used by a user for searching.
A sentence embedder maps each keyword $k$ (or $u$) to an L2-normalized vector $\mathbf{e}(k)$ (or $\mathbf{e}(u)$).
The cosine similarity of two keywords is therefore their dot product:

\begin{equation}
s(k,u)=\mathbf{e}(k)\cdot\mathbf{e}(u)\in[-1,1]
\end{equation}

\vspace{1mm} \noindent
Given the aforementioned sets $K$ and $U$, \textbf{semantic soft precision and recall} are defined as follows: each generated keyword is matched to its most similar ground-truth keyword and each ground-truth keyword is matched to its most similar generated keyword:
\begin{align}
\label{eq:sem_p}
    \mathrm{P_\text{sem}} = \frac{1}{n}\sum_{i=1}^{n}\max_{1\le j\le m} s(k_i,u_j),\\
\label{eq:sem_r}
    \mathrm{R_\text{sem}} = \frac{1}{m}\sum_{j=1}^{m}\max_{1\le i\le n} s(k_i,u_j).
\end{align}
Matches are not exclusive: one keyword can be the best match for several others.
Precision penalizes false-positive or erroneous keywords, because each stored dataset keyword must resemble \emph{some} user keyword.
Recall penalizes missed concepts, because each user keyword must be covered by \emph{some} stored dataset keyword.
 
\textbf{Semantic soft-F1}, F1$_{\mathrm{sem}}$, is the harmonic mean of precision and recall:
\begin{equation}
\label{eq:sem_f1}
\mathrm{F1_{sem}} =
\frac{2\, \mathrm{P_\text{sem}}\, \mathrm{R_\text{sem}}}
     {\mathrm{P_\text{sem}} + \mathrm{R_\text{sem}}}
\end{equation}
It is set to $0$ when either keyword set is empty or when $\mathrm{P_{sem}}+\mathrm{R_{sem}}=0$.
F1$_{\text{sem}}$ rewards keyword sets that are both relevant and complete in meaning, not only in surface form.

Later in Section \ref{sec:Experiments}, we will use semantic soft-F1 as an evaluation metric in our experiments.



\subsection{Keyword Generation Strategies}
\label{sec:Strategies}
As an example, consider that $D_z$ on the right of Fig.~\ref{fig:Workflow} is a virtual dataset that was just created by a user using a transformation function $F_b$. We denote the keyword set to be generated for $D_z$ as $\mathbf{k}_z$, the generic text description of the transformation function as $\mathbf{f}_b$, the keyword sets associated with all input datasets (including $D_y$ and maybe more) as $\mathbf{K} = \{ \mathbf{k}_y, \ldots\}$, and the text descriptions associated with all input datasets as $\mathbf{T} = \{ \mathbf{t}_y, \ldots\}$. As illustrated in Fig.~\ref{fig:Methods}, we can consider five different strategies to provide LLMs with the relevant information to generate the set of keywords $\mathbf{k}_z$. Note that in the figure, we remove the indices for $\mathbf{k}, \mathbf{t}, \mathbf{f}, \mathbf{K}$, and $\mathbf{T}$, as they are distinguishable symbols when the current focus $D_z$ is changed. Fig.~\ref{fig:Methods} is generalizable to any new virtual dataset for which a set of keywords needs to be generated.

\vspace{1mm} \noindent
\textbf{Method $\mathbf{M}_\text{t}$} -- \emph{Using Local Text Description $\mathbf{(t)}$ Only}. To generate a set of keywords $\mathbf{(k)}$ for a newly transformed dataset, a keyword generation process is strictly based on the text description $\mathbf{(t)}$ of the dataset. In our DVS, such a text description is stored in the README file accompanying the dataset.
This approach incurs minimal token consumption and has the lowest execution latency. It is fully isolated from the parent nodes in the DVG graph. It cannot benefit from good-quality metadata (e.g., detailed human documentation) nor be impaired by poor-quality metadata (e.g., incorrectly assigned keywords) at the parent nodes. 

\vspace{1mm} \noindent
\textbf{Method $\mathbf{M}_\text{tf}$} -- \emph{Using $\mathbf{(t)}$ and the Text Description of the Proceeding Transformation Function $\mathbf{(f)}$}. As described in the previous section, users create each virtual data through a UI, which makes related API calls to transform the input data into the desired output data. The UI of each data transformation function (e.g., \texttt{Data Partition}) has some \emph{generic documentation} about the function. This text is denoted as $\mathbf{(f)}$.
It is necessary to note that such generic documentation does not change when the corresponding UI is activated. It does not contain specific information about the datasets that were used as inputs and the values that were set for individual parameters. In our DVS, information about every instance of UI use is stored in the README file of each output dataset.

For example, if a user uses the transformation function \texttt{Data Partition} to divide a dataset \texttt{WholeX} into virtual datasets, \texttt{Xtrn}, \texttt{Xvld}, and \texttt{Xtst}, at the ratio $30:35:35$ for training, validation, and testing. The text of $\mathbf{(f)}$ includes generic descriptions about what \texttt{Data Partition} does and what parameters can be set. Meanwhile, the actual name of the input dataset, \texttt{WholeX}, and the parameter settings (including the ratio 30:35:35) are pushed by the UI into the README files of each \texttt{Xtrn}, \texttt{Xvld}, and \texttt{Xtst}. In other words, these files are the $\mathbf{(t)}$ texts of the three newly created virtual datasets.

\vspace{1mm} \noindent
\textbf{Method $\mathbf{M}_\text{tfK}$} -- \emph{Using $\mathbf{(t)}$, $\mathbf{(f)}$, and the Keyword Sets of All Parent Datasets $\mathbf{(K)}$}. 
With this approach, the keyword generation process traverses the DV graph backward to locate the keyword set assigned to each parent dataset. 
Since there can be more than one parent dataset (e.g., in the case of a merge), these keyword sets, which are collectively denoted as $\mathbf{(K)}$, are used in conjunction with $\mathbf{(t)}$ and $\mathbf{(f)}$. In this way, the process preserves some semantics associated with the lineage of datasets, especially when the same strategy was used to generate the keyword sets of the parent datasets, and so forth. However, if there were errors in the parent keyword sets, they could be propagated to the downstream datasets.

\vspace{1mm} \noindent
\textbf{Method $\mathbf{M}_\text{tfT}$} -- \emph{Using $\mathbf{(t)}$, $\mathbf{(f)}$, and the Text Descriptions of All Parent Datasets $\mathbf{(T)}$}. Instead of using the keyword sets of all parent datasets, a keyword generation process could use the raw text descriptions of the parent datasets instead (i.e., the README files in our DVS). Collectively, these text descriptions are denoted as $\mathbf{(T)}$. In this way, poor-quality keywords generated in previous processes do not enter the new keyword generation process, although LLMs can still make the same mistake with a different collection of texts. Let $\mathbf{(T)}$ contain $m$ text descriptions for parent datasets $m$, the keyword generation process would require more computational cost, as token usage may scale quadratically with $m+2$ text files (i.e., text files of the transformation function $\mathbf{(f)}$, its output dataset $\mathbf{(t)}$, and the $m$ input datasets $\mathbf{(T)}$.

\vspace{1mm} \noindent
\textbf{Method $\mathbf{M}_\text{tfKT}$} -- \emph{Using $\mathbf{(t)}$, $\mathbf{(f)}$, and the Keyword Sets as well as Text Descriptions of All Parent Datasets $\mathbf{(K, T)}$}. This strategy aims to maximize the information available for the keyword generation process. Although the information in $\mathbf{(K)}$ appears to be redundant as it was derived from $\mathbf{(T)}$, $\mathbf{(K)}$ may contain information on the transformation functions and grandparent datasets that were used to generate the parent datasets, unless only the method $\mathbf{M}_\text{t}$ was used previously. $\mathbf{(K)}$ preserves more semantics associated with the lineage of datasets. However, this risks giving too much attention to previous keywords and passing on poor-quality keywords.

\section{Experimental Research}
\label{sec:Experiments}
In the two previous sections, we present the system aspects of this work, delineating the first two contributions outlined in Section \ref{sec:Intro}.  In this section, we focus on the third contribution, reporting a series of experiments designed to answer the following research questions:

\begin{enumerate}[label=\textbf{RQ\arabic*.}, labelwidth=2.5em, leftmargin=3em]
    \item Can LLMs automatically generate relevant keywords from the metadata and user comments stored with each dataset?
    \item Does the quality of generated keywords depend on application domains?
    \item How much does users' effort in writing comments affect the quality of generated keywords?
    \item How much does the users' expertise level affect the quality of generated keywords?
    \item Does utilization of the metadata stored in the system affect the quality of generated keywords?
    \item Which keyword generation strategy is most effective?
    \item Which LLM semantic embedder is more effective?
    \item What LLM sampling temperature is most effective?
    \item How does semantic similarity to ground-truth keywords relate to retrieval performance of LLM-generated keywords in dataset search?
    %
\end{enumerate}

\subsection{Datasets}
\label{sec:Dataset}

We evaluated our keyword generation service (KGS) with datasets in five application domains as reported initially in conjunction with an early version of the data virtualization service (without KGS) ~\cite{Khan:2025:SCC}. These five applications are:

\begin{enumerate}[label=\arabic*), leftmargin=*]
    \item [1,2)] training models for approximating numerical simulation of the Helmholtz and 2D Wave partial differential equations~\cite{Farea:2025:CPC,Farea:HiPC2025};
    \item [3)] training classification and segmentation models for early fault detection in conjunction with the SCADA wind-turbine measurements (multivariate time series)~\cite{gueck2024:scada};
    \item [4)] training speech processing models that are noisy and tolerant, in conjunction with a speech dataset and a noise dataset (audio data)~\cite{valentini2017:speech};
    \item [5)] training multi-task models with multi-variate and multi-label synthetic images of shapes (image data) ~\cite{khan2025:synthetic}; 
\end{enumerate}

Together, there are a variety of datasets, including symbolic, numerical, imagery, tabular, audio, and time series modalities. This allowed us to assess how keyword quality varies among the application domains. 

For each application, we consider datasets constructed in a workflow with data transformations from raw datasets to various virtual datasets required by machine learning tasks and intermediate virtual datasets. 

For each source dataset, we use the original document as its README file. For each virtual dataset, we ask annotators to use the UI (as in Fig. \ref{fig:UI}) to specify it with the appropriate transformation function and parameters together with their comments. The README file generated for each virtual dataset contains two types of data: user comments and (M) the metadata automatically generated by the UI. The annotators were divided into two groups: (E) machine learning experts familiar with individual workflows, and (N) non-experts unfamiliar with the workflows. Each annotator wrote two versions of comments: (L) long and (S) short. Expert annotators also created ground-truth (GT) keywords for the README file of each generated dataset.

Hence, for each virtual dataset, there are nine variants as shown in the first part of Table~\ref{tab:variants}. In addition, we also experimented with partial user comments and partial metadata as listed in the second and third parts of Table~\ref{tab:variants}. 

\begin{table}[t]
\centering
\caption{
Annotation variant codes. Meta.\ cov.\ = fraction of workflow metadata included; Cmt.\ cov.\ = fraction of entries with a comment; Exp.\ = annotator expertise; Len.\ = comment length. 
}
\label{tab:variants}
\setlength{\tabcolsep}{2.5pt}
\footnotesize
\begin{tabular}{@{}lccccl@{}}
    \toprule
    Code & \% Meta. & Exp. & Len. & \% Cmt. & Remark \\
    \midrule
    M      & 100\% & —  & —  &   0\% & metadata-only \\
    EL     &   0\% & E  & L  & 100\% & comment-only, expert, long\\
    ES     &   0\% & E  & S  & 100\% & comment-only, expert, short \\
    NL     &   0\% & N  & L  & 100\% & comment-only, non-expert, long\\
    NS     &   0\% & N  & S  & 100\% & comment-only, non-expert, short\\
    MEL    & 100\% & E  & L  & 100\% & full README, expert, long\\
    MES    & 100\% & E  & S  & 100\% & full README, expert, short\\
    MNL    & 100\% & N  & L  & 100\% & full README, non-expert, long\\
    MNS    & 100\% & N  & S  & 100\% & full README, non-expert, short\\
    \addlinespace[2pt]
    \multicolumn{6}{l}{\textit{Varying comment coverage (partial comment, full metadata):}} \\
    MEL10  & 100\% & E  & L  &  10\% & 10\% user comments\\
    MEL25  & 100\% & E  & L  &  25\% & 25\% user comments\\
    MEL50  & 100\% & E  & L  &  50\% & 50\% user comments\\
    MEL75  & 100\% & E  & L  &  75\% & 75\% user comments\\
    \bottomrule
\end{tabular}
\end{table}

\subsection{Evaluation}
\label{sec:evaluation}
Similar to the scenarios of matching users' keywords with stored dataset keywords, paraphrased keywords (e.g., ``offshore wind'' vs ``wind farm'', ``train/test partition'' vs ``data split'') are not expected to match exactly as text strings. We use the semantic similarity measures as described in Section \ref{sec:Techniques}.

Let $G=\{g_1,\dots,g_n\}$ be the $n$ generated keywords and $T=\{t_1,\dots,t_m\}$ the $m$ ground-truth keywords of a dataset. We use the same semantic metrics $\mathrm{P_\text{sem}}$, $\mathrm{R_\text{sem}}$, and $\mathrm{F1_\text{sem}}$ as defined in Eqs. \ref{eq:sem_p}, \ref{eq:sem_r}, and \ref{eq:sem_f1}.

\vspace{1mm} \noindent 
\textbf{Shuffled ground truth (SGT)} 
is a \emph{null} baseline. It estimates the $\mathrm{F1_\text{sem}}$ obtained when keywords are scored against the ground truth of the wrong dataset \cite{zhang2020:ICLR:bertscore}.
Let $G_i$ be the LLM-generated keywords for dataset $i$, $T_j$ the ground-truth keywords for dataset $j$, and $\mathcal{D}$ the set of all datasets.
We compare $G_i$ with the ground truth of other datasets and average the scores.
We define two variants:

\begin{equation}
\text{SGT-A}_i = \frac{1}{|\mathcal{D}|-1}\sum_{j\in\mathcal{D},\, j\neq i} \mathrm{F1_\text{sem}}(G_i, T_j),
\label{eq:sgt_a}
\end{equation}

\begin{equation}
\text{SGT-W}_i = \frac{1}{|\mathcal{W}_i|-1}\sum_{j\in\mathcal{W}_i,\, j\neq i} \mathrm{F1_\text{sem}}(G_i, T_j),
\label{eq:sgt_w}
\end{equation}
where $\mathcal{W}_i \subseteq \mathcal{D}$ is the set of datasets in the same workflow as dataset $i$, including $i$ itself.
SGT-A (\emph{any workflow}) uses all other datasets and estimates the score expected by chance.
SGT-W (\emph{same workflow}) uses only datasets from the same workflow (e.g., a Windfarm dataset is compared only with other Windfarm datasets).
It measures how much of the score comes from shared domain vocabulary.
We report both as means over all datasets.

\vspace{1mm} \noindent 
\textbf{Rescaled F1} 
\cite{zhang2020:ICLR:bertscore} adjusts F1$_{\text{sem}}$ for the similarity expected by chance.
For each dataset $i$, we use SGT-A$_i$ as the null score:

\begin{equation}
\mathrm{F1}_{\mathrm{rescaled},i} = \frac{\mathrm{F1}_{\text{sem},i} - \text{SGT-A}_i}{1 - \text{SGT-A}_i},
\label{eq:rescaled_f1}
\end{equation}
and we average the result over datasets.
A score of 0 means chance-level performance, and 1 means a perfect semantic match with the ground truth.
Negative scores indicate performance below chance.

\vspace{1mm} \noindent 
\textbf{Jaccard (strict).}
After lower-case and whitespace normalization, the exact-overlap score is measured as

\begin{equation}
J = \frac{|G \cap T|}{|G \cup T|}
\end{equation}
is a hard lower bound that ignores synonymy; the gap between $\mathrm{J}$ and Semantic $\mathrm{F1}$ measures how much paraphrasing the generator introduces.

\vspace{1mm} \noindent 
\textbf{Retrieval metrics.}
For each query $q$, we rank datasets by match score and treat as relevant the set $R_q$ of datasets whose ground-truth keywords contain every query keyword. Let $T_q^{K}$ denote the top $K$ ranked datasets. Recall at $K$ is the fraction of relevant datasets that appear among the top $K$ ranks:

\begin{equation}
  \mathrm{R@}K = \frac{|R_q \cap T_q^{K}|}{|R_q|}.
  \label{eq:recall_at_k}
\end{equation}

R-precision is the precision within the top $|R_q|$ ranks:

\begin{equation}
  \mathrm{RP} = \frac{|R_q \cap T_q^{|R_q|}|}{|R_q|}.
  \label{eq:r_precision}
\end{equation}
When a query has a single relevant node, R-precision is 1 if that node is ranked
first and 0 otherwise. We report both metrics averaged over all queries, with $K = 5$.

\subsection{Results}
\label{sec:Results}
We summarize the results of our experiments according to the nine research questions outlined at the beginning of this section. For each research question, we first state the key finding, then describe the corresponding experiment and results obtained. 


\vspace{1mm} \noindent 
\textbf{RQ1. LLMs can automatically generate relevant keywords from the metadata and user comments stored with each dataset.}

\begin{figure*}[t]
  \centering
  \includegraphics[width=\textwidth, trim=.2cm .3cm .2cm .1cm, clip]{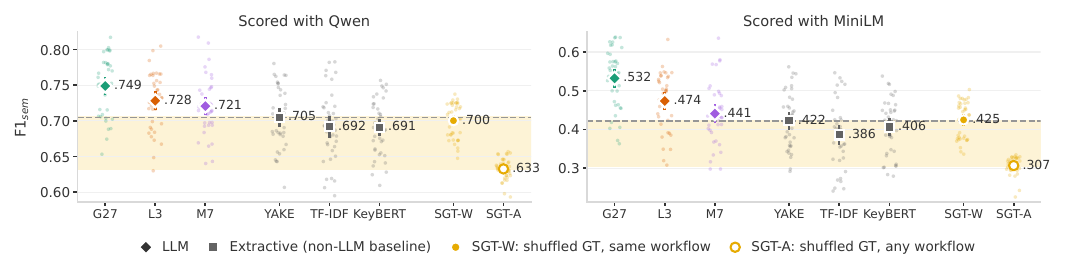}
  \caption{For \textbf{RQ1, RQ7}. 
  $\mathrm{F1_{sem}}$ of LLM keywords, extractive baselines, and shuffled ground truth (SGT), evaluated with Qwen and MiniLM. Each dataset's F1 is averaged over seven input dataset variants and all five methods, at temperature 0.  Extractive baselines do not depend on the method, so they are averaged over the variants only. Large markers show means with 95\% confidence intervals (CIs); small dots represent individual datasets. The yellow band ranges from SGT-A (chance) to the best extractive baseline (dashed line).}
  \label{fig:f1_rq1_floors}
\end{figure*}

\begin{table}[t]
  \caption{
    For \textbf{RQ1, RQ7}. Raw and rescaled keyword-quality scores.
  }
  \label{tab:rq1_rescaled}
  \centering
   \setlength{\tabcolsep}{3.5pt}
  \begin{tabular}{@{}lcccc@{}}
    \toprule
    & \multicolumn{2}{c}{$\mathrm{F1_{sem}}$} & \multicolumn{2}{c}{$\mathrm{F1}_{\mathrm{rescaled}}$} \\
    \cmidrule(lr){2-3}\cmidrule(lr){4-5}
    & Qwen & MiniLM & Qwen & MiniLM \\
    \midrule
    G27                    & .749 & .532 & .30 & .29 \\
    L3                     & .728 & .474 & .27 & .25 \\
    M7                     & .721 & .441 & .23 & .21 \\
    Best extractive$^{a}$  & .705 & .422 & .19 & .16 \\
    SGT-W                  & .700 & .425 & .18 & .17 \\
    SGT-A                  & .633 & .307 & 0 & 0 \\
    \bottomrule
  \end{tabular}\\[2pt]
  \raggedright\scriptsize $^{a}$YAKE for both embedders.
\end{table}

$\mathrm{F1_{sem}}$ has no natural zero because even unrelated scientific terms can have high similarity scores~\cite{zhang2020:ICLR:bertscore}. 
We therefore compare the LLM keywords with two reference points: extractive baselines and shuffled ground truth. 

The \emph{extractive baselines} are non-LLM methods that copy relevant words, phrases, or sentences from the original comment rather than generating new text or keywords. YAKE~\cite{campos2020:InfoSc:yake} ranks phrases using features, e.g., frequency and position; TF-IDF~\cite{sparck1972:TF-IDF} identifies phrases that are common in one README but rare across others; and KeyBERT~\cite{grootendorst2020/:keybert} selects phrases semantically similar to the full README. Each method returns $k=10$ keywords containing 1--3 words. 

The \emph{shuffled ground truth (SGT)} (Section~\ref{sec:evaluation}), is a \emph{null} baseline that estimates $\mathrm{F1_{sem}}$ when reference answers are randomly mismatched with inputs. For each dataset, we compare its LLM-generated keywords with the ground-truth keywords of every other dataset and average the scores.

We evaluated all keyword sets against the ground truth using Qwen and MiniLM embeddings. Fig.~\ref{fig:f1_rq1_floors} and Table~\ref{tab:rq1_rescaled} report results for seven variants (M, MEL, MES, MNL, MNS, EL, and ES), five methods, temperature 0, and 38 datasets from five workflows. We compare each LLM with the best extractive baseline. The table reports raw $\mathrm{F1_{sem}}$ and $\mathrm{F1}_{\mathrm{rescaled}}$, calculated using Eq.~\ref{eq:rescaled_f1}, where 0 represents chance performance. Before averaging across datasets, we rescale each dataset using the SGT-A score for the same keywords. For the extractive baselines and SGT-W, SGT-A is averaged over the three generators. SGT-W uses shuffled ground truth from the same workflow, whereas SGT-A uses shuffled ground truth from any workflow.

For each dataset, we test whether its LLM-generated keywords match its own ground truth better than the ground truth of other datasets (on average). This holds for 95--100\% of datasets when compared with datasets from any workflow, and for 66--79\% when compared only with datasets from the same workflow. These ranges cover the three LLMs and two embedders. Thus, the keywords reliably capture the topic of a dataset and usually distinguish it from related datasets.

G27 performs best: $\mathrm{F1_{sem}}$ is 0.749 with Qwen and 0.532 with MiniLM, vs. 0.705 for YAKE with both embedders. It outperforms the best baseline on 89\% and 92\% of datasets, respectively. Smaller models show limited gains: L3 scores 0.728 and 0.474, M7 scores 0.721 and 0.441, and are not significantly better than extraction.

\begin{figure}[t]
  \centering
  \includegraphics[width=\linewidth, trim=.1cm .3cm .1cm .1cm, clip]{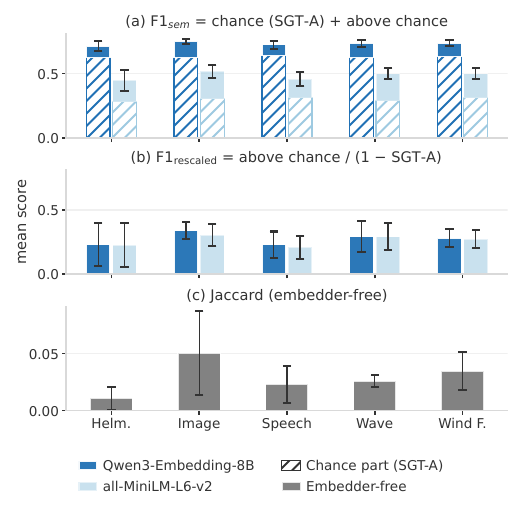}
  \caption{
    For \textbf{RQ7, RQ1}. Embedder comparison per workflow. 
    (a)~Raw $\mathrm{F1_{sem}}$ under Qwen and MiniLM. Each bar is split into the chance part (hatched; SGT-A, RQ1) and the part above chance (solid).
    (b)~$\mathrm{F1}_{\mathrm{rescaled}}$, which divides the part above chance by $1 - \text{SGT-A}$. 
    (c)~Embedder-free Jaccard. 
    Bars show the mean with 95\% CI across nodes. Scores are averaged over the seven input variants, all models, all methods, and temperature 0 and repeats.
  }
  \label{fig:rq7-embedder-metrics}
\end{figure}

After rescaling (Table~\ref{tab:rq1_rescaled} and Fig.~\ref{fig:rq7-embedder-metrics}(b)), Qwen and MiniLM produce similar results. Their respective scores are 0.30 and 0.29 for G27, 0.27 and 0.25 for L3, 0.23 and 0.21 for M7, and 0.19 and 0.16 for the best extractive baseline. The same-workflow null baseline scores 0.18 and 0.17.

A rescaled score of 0.29 does not mean that only 29\% of the keywords are correct. A score of 1 requires every keyword to match the ground truth exactly in meaning; valid synonyms and alternative expert annotations may score lower.

Overall, LLMs generate keywords that match the correct datasets better than chance and outperform phrases extracted from the same inputs. However, the benefit depends on model size: G27 improves on extraction, whereas the smaller models perform similarly to simpler non-LLM methods.


\vspace{1mm} \noindent 
\textbf{RQ2. The quality of generated keywords does not have significant variations among the five application workflows tested, and likely it does not depend on application domains in general.}

\begin{table}[t]
\caption{
  For \textbf{RQ2}. Workflow-level $\mathrm{F1_{sem}}$ scores.
}
\label{tab:rq2_domain}
\centering
\setlength{\tabcolsep}{4pt}
\begin{tabular}{lrrrrr}
\toprule
Workflow & G27 & L3 & M7 & Best extr. & SGT-W \\
\midrule
Helmholtz & \textbf{0.728} & 0.706 & 0.704 & 0.686 & 0.694 \\
Image     & \textbf{0.777} & 0.738 & 0.741 & 0.729 & 0.717 \\
Speech    & \textbf{0.736} & 0.724 & 0.713 & 0.699 & 0.674 \\
Wave      & \textbf{0.753} & 0.732 & 0.722 & 0.691 & 0.720 \\
Windfarm  & \textbf{0.754} & 0.736 & 0.723 & 0.710 & 0.723 \\
\bottomrule
\end{tabular}
\end{table}

Table~\ref{tab:rq2_domain} shows $\mathrm{F1_{sem}}$ for each of five workflows using  Qwen embedder at temperature 0, averaged over seven input variants and five methods, consistent with the settings used for RQ1. ``Best extr.'' denotes the best extractive baseline among YAKE, TF-IDF, and KeyBERT. SGT-W is the shuffled ground-truth baseline, obtained by scoring each dataset's keywords against the ground truth of other datasets in the same workflow or domain. 
Scores are similar across workflows. G27 scores range between 0.777 (Image) and 0.728 (Helmholtz). 

Workflows differ in how much vocabulary their nodes share. SGT-W scores each dataset's keywords against the ground truth of other datasets in the same workflow.
A high SGT-W means that datasets in that workflow use similar words. 
SGT-W scores range between 0.723 (Windfarm) and 0.694 (Helmholtz). Part of the $\mathrm{F1_{sem}}$ comes from shared domain vocabulary, not only from dataset-specific keywords.

Keyword quality does not depend strongly on the domain in our data. G27 performs best and beats extraction in every workflow. Small models add less beyond shared domain vocabulary, and this varies by workflow.


\vspace{1mm} \noindent 
\textbf{RQ3. Users' effort in writing comments does affect the quality of generated keywords.}

\begin{figure*}[t]
    \centering
    \includegraphics[width=\linewidth, trim=.1cm .3cm .1cm .1cm, clip]{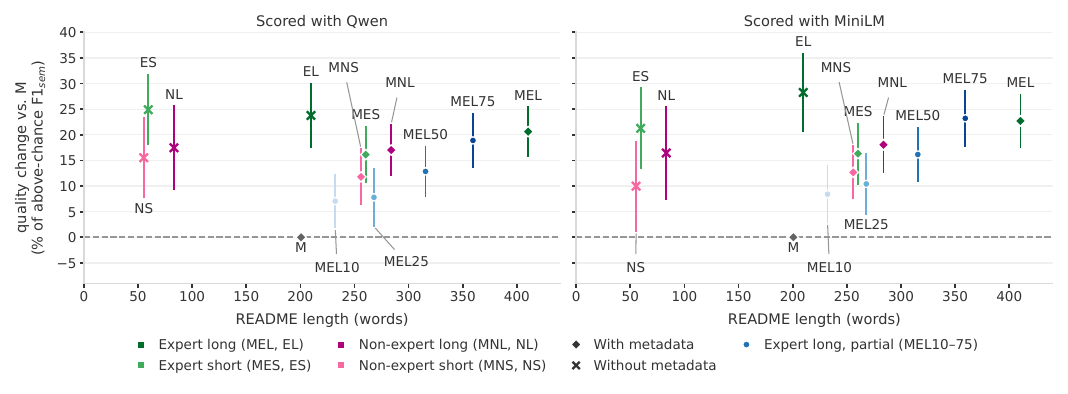}
    \caption{
    For \textbf{RQ3, RQ4, RQ5}. The relative quality statistics of each experimental condition, with the mean value of condition M (metadata only) as the reference point. 
    Keywords were generated at temperature 0 by three generators with all five methods, averaged over all runs.
    The $x$-axis encodes the length of input data, and the $y$-axis encodes the relative changes in percentage against M.
    Error bars show the 95\% CIs of the paired difference to M across the 38 datasets.
    }
    \label{fig:rq3_effort}
\end{figure*}

Fig.~\ref{fig:rq3_effort} shows the quality changes relative to the baseline mean score of the M condition (i.e., metadata only). Given an experiment X (X = M, EL, ES, $\dots$), we measure the quality change as:
\begin{equation}\label{eq:Change}
    \text{change} = \frac{\mathrm{F1_\text{sem}(X)} - \mathrm{F1_\text{sem}(M)}}{\mathrm{F1_\text{sem}(M)} - \text{SGT-A}}
    100 \%
\end{equation}
where SGT-A is the shuffled ground truth as described in Section \ref{sec:evaluation}. With these relative changes, we can observe:

\begin{itemize}[leftmargin=1em, labelsep=0.3em]
    \item EL (expert long, dark green) yields better quality keywords than ES (expert short, light green) with the MiniLM embedder, while they have similar quality if measured with the Qwen embedder.
    \item NL (non-expert long, dark pink) yields better quality keywords than NS (non-expert short, light pink) with both embedders.
    \item When metadata is used with user comments, the mean scores show longer comments are better than shorter comments, i.e., MEL $>$ MES and MNL $>$ MNS.
    \item When we reduce the length of expert comments from 100\% to 75\%, 50\%, 25\%, 10\%, and 0\%, we can see the strong correlation with the quality reduction of the mean values of MEL (green), MEL75 (dark blue), MEL50 (blue), MEL25 (light blue), MEL10 (sky blue), and M (gray).
    
\end{itemize}


\vspace{1mm} \noindent 
\textbf{RQ4. Users' expertise level does affect the quality of generated keywords.}

From Fig.~\ref{fig:rq3_effort}, we can also compare the comments by experts and non-experts. We can observe:
\begin{itemize}[leftmargin=1em, labelsep=0.3em]
    \item Regardless of with or without metadata, long or short comments, and the use of either embedder, the mean values consistently show that expert comments yield better quality keywords than non-expert comments, i.e., EL $>$ NL, ES $>$ NS, MEL $>$ MNL, and MES $>$ MNS.
    \item The numerical differences between the mean values are:
    \begin{itemize}
        \item NL vs. EL: $-6.3\%$ (Qwen) / $-11.8\%$ (MiniLM),
        \item NS vs. ES: $-9.4\%$ (Qwen) / $-11.3\%$ (MiniLM),
        \item MNL vs. MEL: $-3.6\%$ (Qwen) / $-4.6\%$ (MiniLM),
        \item MNS vs. MES: $-4.3\%$ (Qwen) / $-3.7\%$ (MiniLM).
    \end{itemize}
    \item The gaps are narrower when metadata is used in conjunction with user comments. We will discuss this further in the context of \textbf{RQ5}.
\end{itemize}


\vspace{1mm} \noindent 
\textbf{RQ5. The effectiveness of utilizing the metadata stored in the DVS depends on the availability and expertise of user comments.}

\begin{table}[t]
\centering
\caption{
    For \textbf{RQ5}. Keyword quality with metadata and expert comment.
    }
\label{tab:rq5_metadata_only}
\begin{tabular}{lcc}
    \toprule
     & Qwen & MiniLM \\
    \midrule
    
    a. SGT-A & .633 & .307 \\
    b. $\mathrm{F1_\text{sem}}$(M) & .719 & .458 \\
    c. $\mathrm{F1_\text{sem}}$(MEL) & .736 & .491 \\
    \midrule
    (b -- a) & +.086 & +.151 \\
    (c -- b) & +.017 & +.033 \\
    \bottomrule
\end{tabular}
\end{table}

From Fig.~\ref{fig:rq3_effort}, in terms of mean values, we can observe:

\begin{itemize}[leftmargin=1em, labelsep=0.3em]
    \item EL $>$ MEL and ES $>$ MES  (with both embedders).
    \item NL $\sim$ MNL (with both embedders).
    \item NS $\sim$ MNS (small differences of opposite sign (Qwen $+$, MiniLM $-$)).
\end{itemize}

This suggests that the utilization of metadata is not effective when expert comments are available. However, with a non-expert comment, metadata has no measurable effect.

Table~\ref{tab:rq5_metadata_only} reports the mean $\mathrm{F1_\text{sem}}$ over 38 datasets from five workflows at temperature 0, averaged over three models, five methods, and all repeats. SGT-A is the random baseline, obtained by scoring the keywords of seven input variants (M, MEL, MES, MNL, MNS, EL, and ES) against the ground truth of other datasets (see RQ1). 
The difference (b--a) represents the gain achieved by adding metadata to SGT-A: +0.086 for Qwen and +0.151 for MiniLM. The difference (c--b) represents the additional gain from incorporating an expert comment: +0.017 for Qwen and +0.033 for MiniLM.
Meanwhile, metadata remains useful when no comment is available.

If we plot the SGT-A scores in Fig. \ref{fig:rq3_effort} using Eq.\,\ref{eq:Change}, they would appear at $(0, -100\%)$ in both the left and right plots. In comparison, the gaps between M and MEL are:
21\% (Qwen) and 23\% (MiniLM). The gaps between M and SGT-A are more than 4 times as large as the gaps between MEL and M.


\vspace{1mm} \noindent 
\textbf{RQ6. The keyword generation strategies that utilize parents' keywords are most effective.}


\begin{figure}[t]
    \centering
    \includegraphics[width=\linewidth, trim=.1cm .3cm .1cm .1cm, clip]{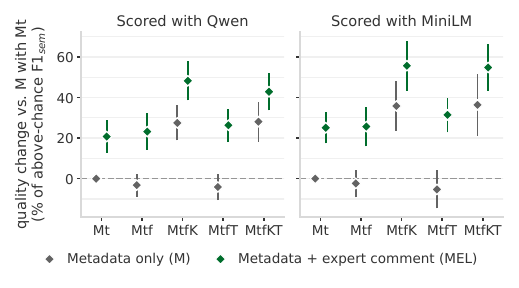}
    \caption{For \textbf{RQ6}. The keyword generation of the current dataset (child) may benefit from the information stored in the parent datasets. The experiments were conducted under two conditions: (a, M) the child dataset has no user comments and (b, MEL) has long comments from an expert user.}
    \label{fig:rq6_methods_mel_vs_m}
\end{figure}

\begin{table}[t]
\centering
\caption{For \textbf{RQ6}. Results for five keyword-generation strategies under the M and MEL conditions.
}
\label{tab:rq6_methods}
\setlength{\tabcolsep}{3pt}
\begin{tabular}{@{}lc|lccccc@{}}
    \toprule
     embedder & condition & & Mt & Mtf & MtfK & MtfT & MtfKT \\
    \midrule
    \multirow{4}{*}{Qwen} & \multirow{2}{*}{(a, M)} & $\mathrm{F1_{sem}}$ & .712 & .709 & .733 & .708 & .733 \\
     & & change\% & --  & $-$3.2 & 27.5 & $-$4.1 & 28.1 \\
    \cmidrule(lr){2-8}
     & \multirow{2}{*}{(b, MEL)} & $\mathrm{F1_{sem}}$ & .727 & .729 & .748 & .732 & .744 \\
     & & change\% & 20.8 & 23.2 & 48.2 & 26.3 & 42.8 \\
    \midrule
    \multirow{4}{*}{MiniLM} & \multirow{2}{*}{(a, M)} & $\mathrm{F1_{sem}}$ & .441 & .438 & .487 & .434 & .488 \\
     & & change\% & --  & $-$2.4 & 35.8 & $-$5.3 & 36.3 \\
    \cmidrule(lr){2-8}
     & \multirow{2}{*}{(b, MEL)} & $\mathrm{F1_{sem}}$ & .473 & .474 & .513 & .481 & .511 \\
     & & change\% & 25.1 & 25.7 & 55.7 & 31.4 & 54.8 \\
    \bottomrule
\end{tabular}
\end{table}

Fig.~\ref{fig:rq6_methods_mel_vs_m} and Table \ref{tab:rq6_methods} compare the five strategies described in Section \ref{sec:Strategies}. The main hypothesis that led to the development of the five strategies was that keyword generation for a specific dataset (referred to as \emph{child}) may benefit from the information that is available in the DVS but not in the current README file (t). Such information includes:

\begin{itemize}[leftmargin=1em, labelsep=0.3em]
    \item (f) --- the description of the transformation function used to create the child dataset.
    \item (K) --- the keywords of the \emph{parent} dataset(s).
    \item (T) --- the README file(s) of the \emph{parent} dataset(s).
\end{itemize}

We therefore conducted experiments to evaluate the five strategies, Mt, Mtf, MtfK, MtfT, and MtfKT. In addition to the two embedders, we also consider two scenarios: (a, M) the user who created the child dataset did not provide any comment, and (b, MEL) the user who created the child dataset provided long comments as an expert.  

With the raw $\mathrm{F1}_\text{sem}$ scores, as shown in Table \ref{tab:rq6_methods}, under the condition (a, M), MtfK and MtfKT have made improvements over Mt, while Mtf and MtfT are ineffective. Under the condition (b, MEL), Mtf, MtfK, MtfT, and MtfKT all improve upon Mt, while MtfK and MtfKT are noticeably more effective. The results strongly suggest that using keywords from the parent dataset(s) is beneficial and effective.

In both plots in Fig.~\ref{fig:rq6_methods_mel_vs_m}, we can also notice that the third and fifth gray dots are higher than the first and second green dots. This means that MtfK and MtfKT under the condition of (a, M) perform better than Mt and Mtf under the condition (b, MEL). In other words, using keywords of parent datasets can often bring more benefits than the expert long comments in the current child README file.
The values plotted in Fig.~\ref{fig:rq6_methods_mel_vs_m} are the change\% values reported in Table~\ref{tab:rq6_methods}, converted using Eq.\,\ref{eq:Change}. The table also reports the corresponding raw $\mathrm{F1}_\text{sem}$ scores. For each embedder, the change\% values are calculated relative to Mt with metadata only (M) and SGT-A.


\vspace{1mm} \noindent 
\textbf{RQ7. There is no statistically significant difference between the two LLM semantic embedders.}

Fig.~\ref{fig:rq7-embedder-metrics} compares two embedders for five workflows.
(a) Qwen produces consistently higher raw scores than MiniLM, with a mean difference of approximately 0.25 in $\mathrm{F1_{sem}}$ (0.24--0.27 per workflow).
However, this difference is a property of the embedding space, not of the keywords: the chance floor (SGT-A) is also much higher under Qwen (0.62--0.64 vs. 0.28--0.31). 
(b) Once scores are rescaled in relation to chance, the two embedders agree closely in every workflow differences $\leq$0.03). 
(c) The embedder-free Jaccard shows a largely similar workflow-level pattern without any embedding model.

Neither embedder is more effective in a way that matters: embedder choice affects the numerical scale of semantic scores, but the two embedders mostly produce the same ordering when comparing two experimental conditions. Occasionally they produce different orderings, indicating uncertainty in conclusions. For this reason, we report the relative changes of both embedders against the M condition when discussing research questions, while avoiding using raw scores as metrics.


\vspace{1mm} \noindent 
\textbf{RQ8. Varying LLM sampling temperature has relatively small impact, while zero temperature is a relatively optimal setting for keyword quality.}

\begin{figure}[t]
    \centering
    \includegraphics[width=\linewidth, trim=.1cm .3cm .1cm .1cm, clip]{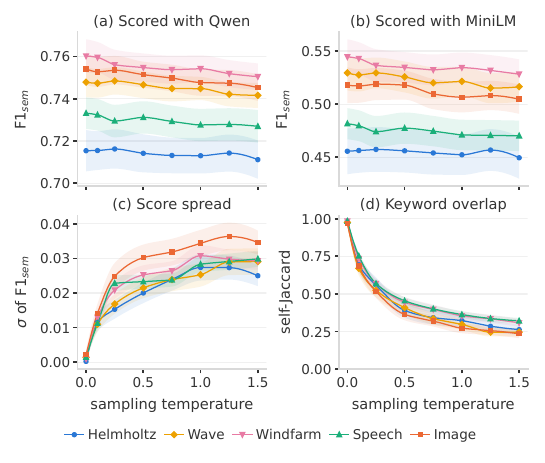}
    \caption{
    For \textbf{RQ8}. 
    Mean $\mathrm{F1_{sem}}$ per workflow vs. sampling temperature (strategy MtfK, five applications, three generators, three runs). 
    $\sigma$: standard deviation of $\mathrm{F1_{sem}}$ across runs.
    Self-Jaccard: mean pairwise overlap of the generated keywords across runs.
    Color bands show 95\% CIs.
    }
    \label{fig:rq8_stability}
\end{figure}

\begin{table}[t]
\centering
\caption{
    For \textbf{RQ8}. Mean $\mathrm{F1_{sem}}$ differences by workflow and temperature.
}
\label{tab:rq8_workflow}
\footnotesize
\setlength{\tabcolsep}{3pt}
\begin{tabular}{lrrrrrrr}
    \toprule
    \hfill T: & 0.00 & 0.25 & 0.50 & 0.75 & 1.00 & 1.25 & 1.50 \\
    \midrule
    \multicolumn{8}{l}{\textit{Scored with Qwen}} \\
    Helmholtz & 0.715 & $+.001$ & $-.001$ & $-.002$ & $-.002$ & $-.001$ & $-.004$ \\
    Wave & 0.748 & $+.001$ & $-.001$ & $-.003$ & $-.003$ & $-.006$ & $-.006$ \\
    Windfarm & 0.760 & $-.004$ & $-.005$ & $-.006$ & $-.006$ & $-.008$ & $-.010$ \\
    Speech & 0.733 & $-.004$ & $-.002$ & $-.004$ & $-.006$ & $-.005$ & $-.006$ \\
    Image & 0.754 & $-.001$ & $-.003$ & $-.004$ & $-.006$ & $-.007$ & $-.009$ \\
    \midrule
    
    \multicolumn{8}{l}{\textit{Scored with MiniLM}} \\
    Helmholtz & 0.456 & $+.002$ & .000 & $-.002$ & $-.003$ & $+.001$ & $-.006$ \\
    Wave & 0.529 & .000 & $-.004$ & $-.009$ & $-.008$ & $-.014$ & $-.013$ \\
    Windfarm & 0.544 & $-.008$ & $-.010$ & $-.012$ & $-.010$ & $-.013$ & $-.016$ \\
    Speech & 0.482 & $-.007$ & $-.004$ & $-.007$ & $-.011$ & $-.011$ & $-.011$ \\
    Image & 0.518 & $+.001$ & .000 & $-.008$ & $-.011$ & $-.010$ & $-.013$ \\
    \midrule
\end{tabular}
\end{table}

Temperature (T) controls the randomness of LLMs in selecting words as keywords. 
Table~\ref{tab:rq8_workflow} reports the mean differences relative to T=0 for strategy MtfK across five applications, three generators, and three runs. The results are plotted in Fig.~\ref{fig:rq8_stability}.

As shown in the lower two plots in Fig.~\ref{fig:rq8_stability}, between T=0 and T=1.5, the higher the temperature, the more diverse the keywords generated (measured using standard deviation) and the less overlap among keywords generated in multiple repeated runs (measured using self-Jaccard). The five lines in each plot represent the experimental results for the five application domains, respectively. 
The top two plots in Fig.~\ref{fig:rq8_stability} show the corresponding keyword quality measured using $\mathrm{F1}_\text{sem}$ with embedders Qwen and MiniLM, respectively.

The results show a general trend of decreasing quality as temperature changes. 
Although small, the measured differences should be converted to relative changes using Eq.~\ref{eq:Change} when interpreted with Fig.~\ref{fig:rq3_effort}. Thus, $+0.001$ represents positive changes of about 1.25\% for Qwen and 0.72\% for MiniLM. At $T=1.5$, quality decreases across all five applications by about 5\%--12.5\% (Qwen) and 4.3\%--11.5\% (MiniLM).

However, we also observe quality gains for some applications when T = 0.25. In other words, for some datasets, keyword generation can benefit from a small increase in temperature. Nevertheless, such gains are not universal. For example, setting T = 0.25 has a positive impact on two applications, but there is a noticeable negative impact on the WindFarm and Speech applications.
We therefore used T = 0 for all main experiments reported in this paper.


\vspace{1mm} \noindent 
\textbf{RQ9. Higher semantic similarity to the ground truth generally means better retrieval across input variants.}

\begin{table}[t]
\centering
\caption{
    For \textbf{RQ9}. Retrieval performance of LLM-generated keywords.
    }
\label{tab:rq9_retrieval}
\setlength{\tabcolsep}{3pt}
\begin{tabular}{lcccc}
    \toprule
     & \multicolumn{2}{c}{R@5} & \multicolumn{2}{c}{R-precision} \\
    \cmidrule(lr){2-3} \cmidrule(lr){4-5}
     & Qwen & MiniLM & Qwen & MiniLM \\
    \midrule
    a. Random ranking & .132 & .132 & .032 & .032 \\
    b. M & .618 & .618 & .240 & .241 \\
    c. MEL, MES, MNS, MNL & .669--.689 & .660--.690 & .272--.290 & .270--.292 \\
    d. EL, ES, NL, NS & .669--.682 & .673--.686 & .259--.269 & .260--.276 \\
    \midrule
    (c -- b) & .051--.071 & .042--.072 & .032--.050 & .029--.051 \\
    (d -- b) & .050--.064 & .055--.068 & .020--.029 & .020--.036 \\
    \bottomrule
\end{tabular}
\end{table}

For each configuration (LLM, input variant, and strategy at temperature 0), we build a \textit{search index} in which each of the 38 datasets is represented by its LLM-generated keywords. We embed all keywords with both embedders (Qwen and MiniLM), so search matches query and dataset keywords by meaning rather than exact words.

We build \textit{search queries} from the ground-truth keywords. Each query contains $k \in [1, 10]$ ground-truth keywords from one dataset that has a keyword set $K$. When $k < |K|$, the corresponding query contains $k$ keywords randomly selected from $K$. When $k \geq |K|$, we choose all keywords in $K$ while resetting $k$ to $|K|$. A dataset is most relevant to a query if its keyword set $K$ contains all $k$ keywords in a query.
So the task of the experiment is to test whether the generated keywords enable the right datasets to be found.

Table~\ref{tab:rq9_retrieval} compares the four retrieval settings using R@5 and R-precision with Qwen and MiniLM. 
Given all datasets available in the search pool, random ranking of these datasets provides the baseline.
In comparison with this baseline, keywords generated from metadata alone (M) already make datasets easy to find, e.g., R@5 is .618, against .132 for a random ranking. Adding a user comment improves search further, raising R@5 by .04--.07 and R-precision by .02--.05. 
For R@5, comments without metadata (EL, ES, NL, NS) help as much as comments with metadata (MEL, MES, MNL, MNS). Metadata adds value mainly at the top of the ranking, where the combined variants reach the highest R-precision (up to .292). 
Both embedders give the same picture. Absolute R-precision stays below .30, so keyword search works best as a short list of candidates rather than a single answer.

\section{Conclusion}
In this paper, we present an enhanced data virtualization service (DVS) with an automated keyword generation service (KGS). By utilizing LLMs, the KGS can save users of the DVS a huge amount of time in specifying keywords, while providing the DVS with an efficient and consistent means for indexing a huge number of datasets. In addition to the use of LLMs, the novelty of this KGS also stems from the utilization of UI-generated metadata as well as users' comments, the utilization of other information available in the DVS through the data virtualization graph, and the design and analysis of five strategies for keyword generation.

To evaluate the KGS, we conducted a series of experiments to investigate nine research questions in conjunction with five applications. While the experiments have provided detailed technical answers to these nine research questions, they have also confirmed the benefits, advantages, and feasibility of the techniques presented in this paper.

In future work, we will introduce a human-in-the-loop facility that presents generated keywords as an editable draft in an interactive user interface and integrates provenance visualization, allowing users to inspect the lineage of discovered datasets. We will also conduct more experiments to measure the quality of text descriptions using different quality metrics, aiming to identify how different types of quality relate to the performance of different keyword generation strategies.

\section*{Acknowledgments}
This work is part of the VIS4ML4HD project funded by UK Research and Innovation (EPSRC: EP/X029557/1).

\bibliographystyle{IEEEtran}
\bibliography{references}


\newpage

\vfill

\end{document}